\documentclass[11pt]{article}

\usepackage[preprint]{acl}
\usepackage{times}
\usepackage{latexsym}
\usepackage{amsmath}
\usepackage{makecell}
\usepackage{amsmath}
\usepackage{amssymb}
\usepackage{graphicx}
\usepackage{booktabs}
\usepackage[T1]{fontenc}

\usepackage[utf8]{inputenc}

\usepackage{microtype}

\usepackage{inconsolata}

\usepackage{graphicx}
 \usepackage{amsmath} 
 \usepackage{amssymb}
 \usepackage{array}

\usepackage{booktabs}
\title{SED-SFT: Selectively Encouraging Diversity in Supervised Fine-Tuning}
\author{ Yijie Chen\thanks{~~Equal contribution.},Yijin Liu\footnotemark[1] and Fandong Meng\\
    WeChat AI, Tencent Inc, China\\
      \{kantichen, yijinliu, fandongmeng\}@tencent.com  }

\begin{document}
\maketitle


\begin{abstract}
Supervised Fine-Tuning (SFT) followed by Reinforcement Learning (RL) has emerged as the standard post-training paradigm for large language models (LLMs). However, the conventional SFT process, driven by Cross-Entropy (CE) loss, often induces mode collapse, where models over-concentrate on specific response patterns. This lack of distributional diversity severely restricts the exploration efficiency required for subsequent RL. While recent studies have attempted to improve SFT by replacing the CE loss, aiming to preserve diversity or refine the update policy, they fail to adequately balance diversity and accuracy, thereby yielding suboptimal performance after RL. To address the mode collapse problem, we propose SED-SFT, which adaptively encourages diversity based on the token exploration space. This framework introduces a selective entropy regularization term with a selective masking mechanism into the optimization objective. Extensive experiments across eight mathematical benchmarks demonstrate that SED-SFT significantly enhances generation diversity with a negligible computational overhead increase compared with CE loss, yielding average improvements of 2.06 and 1.20 points in subsequent RL performance over standard CE-based baselines on Llama-3.2-3B-Instruct and Qwen2.5-Math-7B-Instruct, respectively.~\footnote{The code is publicly available at \url{https://github.com/pppa2019/SED-SFT}}

\end{abstract}

\section{Introduction}

The prevailing post-training paradigm for Large Language Models (LLMs) typically involves Supervised Fine-Tuning (SFT) followed by Reinforcement Learning (RL). SFT primarily serves to align models with human instructions and enhance specialized capabilities cost-effectively. The mainstream research in SFT has focused on data engineering~\cite{gunasekar2023textbooks, lambert2024tulu, zhou2023lima}. However, the standard Cross-Entropy (CE) loss mechanism, which pushes probability mass towards the target label, implicitly suppresses diversity. This reduction in generation diversity~\cite{chen2025retaining, o2024attributing, lin2025debunk} severely constrains the model's exploration space during the subsequent RL phase.
To mitigate mode collapse, existing studies have proposed extending beyond pure CE loss by modifying the update policy or incorporating diversity regularization terms. However, their application is often restricted to the RL stage~\cite{wang2025beyond}, or the performance trade-offs incurred during the SFT phase may not be fully recovered in subsequent RL steps.

Through an analysis of mathematical tasks, we identify the token-level exploration space as a critical factor, and this non-uniformity is a critical factor hindering accuracy when blindly encouraging diversity. 
To this end, we introduce SED-SFT (Selectively Encouraging Diversity in Supervised Fine-Tuning), which incorporates a concise regularization term to modulate the prediction probabilities of target labels, alongside a masking strategy designed to prevent excessive diversity enhancement in regions with limited exploration space.
Our research focuses primarily on mathematical reasoning tasks. We evaluated SED-SFT using two prominent base models across eight rigorous mathematical benchmarks. The results demonstrate significant performance gains, most notably an average improvement of 2.06 points on the Llama-3.2-3B-Instruct model. A detailed analysis of the training process confirms that SED-SFT effectively enhances generative diversity during the SFT stage, acting as a vital catalyst for the superior performance observed in the subsequent RL phase.
\begin{figure*}[ht]
    \centering
    \includegraphics[width=0.9\linewidth]{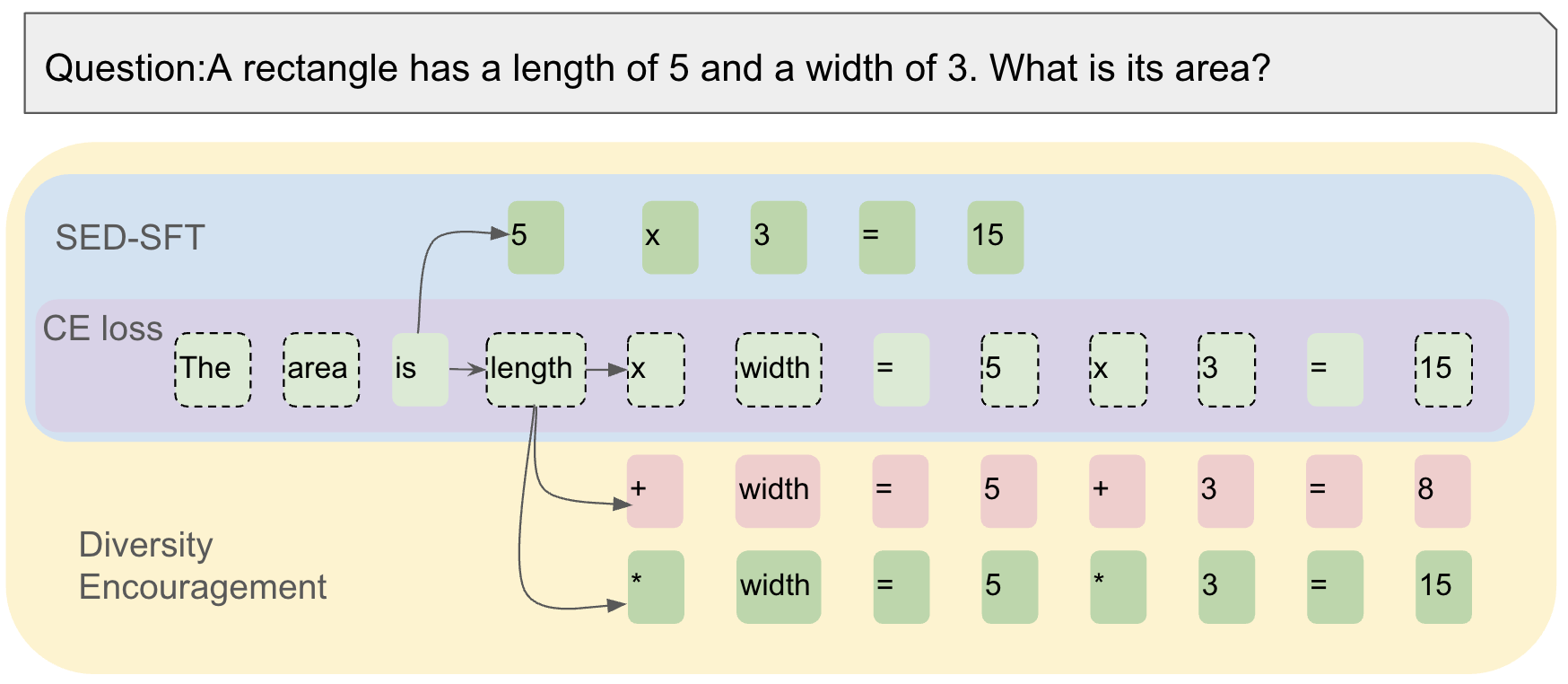}
    \caption{The comparison of Cross-Entropy, pure diversity-encouraging, and SED-SFT. The dashed line token boxes indicate the tokens will be masked in SED-SFT, {\em i.e.}, the tokens with low exploration space. SED-SFT achieves a balance between accuracy and diversity by avoiding encouraging the masked tokens.}
    \label{fig:SED-SFT_concept}
\end{figure*}

Our main contributions are summarized as follows:
\begin{itemize}
\item We identify and analyze the discrepancies within the token-level exploration space in the diversity encouraging process, thereby justifying the necessity of selective updating.
\item We propose SED-SFT, including a concise entropy penalty to foster diversity and a masking strategy to maintain predictive precision.
\item Experiments conducted on two mainstream backbones across eight math benchmarks indicate that SED-SFT significantly improves model performance after the RL process, consistently outperforming the baselines.
\end{itemize}

\section{Preliminary Analysis}
This section provides a systematic analysis of token-level exploration space in SFT. We start by formulating the Cross-Entropy objective, then show why preserving diversity is unnecessary for certain tokens, and finally introduce cumulative top-$k$ probability as a practical metric for quantifying exploration space.
\subsection{On Cross-Entropy in SFT}
SFT trains the model $\pi_{\theta}$ by minimizing the standard CE loss $\mathcal{L}_{\text{CE}}$:
$$\mathcal{L}_{\text{CE}}(\theta) = - \mathbb{E}_{(x, y^*) \sim \mathcal{D}} \left[ \sum_{t=1}^{|y^*|} \log \pi_{\theta}(y^*_t | x, y^*_{<t}) \right]$$
where $y^*$ is the given label sequence, this loss formulation strongly drives the model policy $\pi_{\theta}$ to quickly converge along the single correct path $y^*$, leading to mode collapse and a significant decline in generation diversity~\cite{o2024attributing}.

\subsection{Discrepancies in Token-level Exploration}\label{sec:label_prob_analysis}
By viewing the selection of $y^*$ as a binary event and SFT as a policy update process, the probabilities of ground-truth labels serve as the determining factors for the magnitude of the model update~\cite{zhang2025policy}. To mitigate mode collapse, updates should be strategically restricted to positions that potentially contain alternative reasoning branches.
We analyze the ground-truth label probabilities through a case study (see the visualization heatmap in Appendix~\ref{sec:prob_heatmap} Figure~\ref{fig:heatmap_example}). The results indicate that \textbf{tokens with restricted exploration space}—typically comprising fixed conjunctions, structural tokens, or specific vocabulary—exhibit significantly higher prediction confidence compared to the average prediction score. Consequently, to foster diversity and enrich the model's reasoning paths, preserving diversity on such tokens is unnecessary and may even lead to a degradation in accuracy.

\subsection{Quantifying the Token Exploration Space}
While the analysis in Section~\ref{sec:label_prob_analysis} reveals that a selective strategy during large-scale SFT requires an automated metric to identify tokens with low diversity requirements. To bridge this gap, we seek a quantitative measure that can reflect the exploration space at each position.
Specifically, we evaluate the cumulative top-k probability distribution, a metric that is computationally efficient and robust to semantically equivalent token branches~\cite{kuhn2023semantic}. 
Our analysis of 100 mathematical problems (Figure~\ref{fig:cumsum_divergence}) reveals a trade-off in the choice of $k$. While the median cumulative probability is most discriminative at $k=1$, it offers limited information regarding potential alternative paths. Conversely, as $k$ increases, the metric tends toward uniformity, losing its ability to distinguish restricted tokens from flexible ones. Therefore, we employ $k=2$ or $3$ to characterize the exploration space effectively, and we use this metric to construct the selective mask in the next section.

\section{Selective Diversity Encouragement}

In this section, we introduce SED-SFT, a simple yet effective training objective designed to encourage the model to maintain diversity at appropriate positions (see Figure~\ref{fig:SED-SFT_concept}).

\paragraph{Top-k Masking Strategy $M_t$}
To selectively apply the $\mathcal{L}_{\text{DE}}$, we define a binary mask $M_t$ based on the model's prediction distribution at position $t$. We quantify the token exploration space using the cumulative probability of the Top-$k$ tokens:

$$P_{\text{Top-k}}(t) = \sum_{j \in \mathcal{K}_t} \pi_{\theta}(y_j | x, y^*_{<t})$$

$$ M_t=\mathbf{1}\left[P_{\text{Top-}k}(t)< \tau\right] $$
where $\mathcal{K}_t$ denotes the set of indices for the $k$ tokens with the highest probabilities at step $t$. To obtain a suitable threshold, we define the masking ratio $r$ and let $\mathcal{P} = \{P_{\text{Top-k}}(t)\}_{t=1}^T$ denote the set of cumulative probabilities over all positions in the sampled subset of the training data. We define the threshold $\tau$ as the $(1-r)$-th quantile of $\mathcal{P}$:
$$\tau = \text{Quantile}(\mathcal{P}, 1-r)$$

\paragraph{Diversity Encouraging Function $\mathcal{L}_{\text{DE}}(p)$}
The objective of $\mathcal{L}_{\text{DE}}(p)$ is to suppress the probability $p$ of the correct label by pushing it towards $0.5$, which corresponds to the point of maximum information entropy. We adopt a concise quadratic function, inspired by CHORD~\cite{zhang2025policy}, as the penalty term. This function attains its minimum penalty at $p=0.5$ and maximal penalty at $p=1.0$ or $p=0.0$.
$$\mathcal{L}_{\text{DE}}(p) =  \left(p - \frac{1}{2}\right)^2$$
Here, $p = \pi_{\theta}(y^*_t | x, y^*_{<t})$ is the probability assigned to the ground-truth token $y^*_t$. By minimizing $\mathcal{L}_{\text{DE}}$, $p$ is encouraged to approach $0.5$, thereby allocating more probability mass to alternative plausible paths.

\paragraph{Overall Loss $\mathcal{L}_{\text{SED-SFT}}$}
The final loss combines the diversity encouraging function and the CE loss. Let $\lambda$ be the weight of the diversity encouraging term, and all of our experiments use $\lambda=1$.

$$
\begin{aligned}
\mathcal{L}_{\text{SED-SFT}}(\theta) &= 
\sum_{t=1}^{|y^*|} \Bigl[  -\log \pi_{\theta}(y^*_t \mid x, y^*_{<t}) \\
& + \lambda \cdot M_t \cdot \mathcal{L}_{\text{DE}}
\bigl( \pi_{\theta}(y^*_t \mid x, y^*_{<t}) \bigr) \Bigr]
\end{aligned}
$$

We next validate SED-SFT under the standard SFT-then-RL pipeline and compare it with representative SFT objectives in Section~\ref{tab:main_result}.

\begin{table*}[ht]
\resizebox{\linewidth}{!}{
\begin{tabular}{llrrrrrrrrr}
\toprule
   &
   &
  \multicolumn{3}{c}{Avg@8} &
  \multicolumn{1}{l}{} &
  \multicolumn{1}{l}{} &
  \multicolumn{1}{l}{} &
  \multicolumn{1}{l}{} &
  \multicolumn{1}{l}{} &
  \multicolumn{1}{l}{} \\
  \cline{3-5}
  
 &
   &
  \multicolumn{1}{c}{aime24} &
  \multicolumn{1}{c}{aime25} &
  \multicolumn{1}{c}{amc23} &
  \multicolumn{1}{c}{gsm8k} &
  \multicolumn{1}{c}{math} &
  \multicolumn{1}{c}{gaokao} &
  \multicolumn{1}{c}{olympiad} &
  \multicolumn{1}{c}{college} &
  \multicolumn{1}{c}{average} \\
  \midrule
\multicolumn{11}{c}{Qwen2.5-Math-7B-Instruct}  \\

SFT & CrossEntropy     & 7.10  & 14.20 & 46.90 & 87.70 & 66.80 & 59.50                & 26.10 & 38.30 & 43.33 \\
    & GEM              & 5.00  & 9.20  & 35.30 & 86.80 & 67.70 & 56.60                & 28.00 & 38.50 & 40.89 \\
    & DFT              & 17.10 & 17.50 & 58.10 & 95.30 & 83.90 & 70.10                & 44.00 & 47.90 & 54.24 \\
    & SED-SFT w/o mask & 6.20  & 11.70 & 40.00 & 87.40 & 67.00 & 59.70                & 27.70 & 38.20 & 42.24 \\
    & SED-SFT          & 6.70  & 12.10 & 42.20 & 87.70 & 67.50 & 57.70                & 26.50 & 37.90 & 42.29 \\
    \hline
RL  & CrossEntropy     & 15.40 & 16.20 & \textbf{67.80} & 94.50 & 85.80 & 72.50                & 46.50 & \textbf{49.30} & 56.00 \\
    & GEM              & 12.10 & 16.70 & 63.70 &\textbf{ 96.00} & \textbf{87.00} & 70.60                & 48.90 & 48.70 & 55.46 \\
    & DFT              & 16.70 & 18.30 & 61.60 & \textbf{96.00} & 85.40 & 70.60                & 45.80 & 48.50 & 55.36 \\
    & SED-SFT w/o mask & \textbf{20.00} & 16.70 & 67.20 & 94.90 & 86.00 & 71.40                & 48.30 & 48.30 & 56.60 \\
    & SED-SFT          & 18.80 & \textbf{18.80} & 66.20 & 95.20 & 86.60 & \textbf{73.00}                & \textbf{50.20} & 48.80 & \textbf{57.20} \\
\hline
  \multicolumn{11}{c}{Llama-3.2-3B-Instruct} \\
SFT & CrossEntropy     & 0.80  & 1.20  & 17.20 & 60.70 & 34.10 & 34.50                & 9.60  & 19.90 & 22.25 \\
    & GEM              & 0.00  & 1.20  & 10.30 & 59.20 & 31.60 & 33.80                & 9.90  & 18.60 & 20.58 \\
   
    & DFT              & 2.10   & 0.80   & 16.60  & 69.90  & 39.50  & 37.10          	  & 12.10 &	25.30 &25.43 \\
    & SED-SFT w/o mask & 0.00  & 1.20  & 12.80 & 60.10 & 34.10 & 33.00                & 10.70 & 19.90 & 21.48 \\
    & SED-SFT          & 0.80  & 1.20  & 16.20 & 61.20 & 34.30 & 32.70                & 9.60  & 20.30 & 22.04 \\
    \hline
RL  & CrossEntropy     & 9.20  & 0.00  & 33.40 & 83.50 & 56.40 & 47.80                & 20.90 & 34.00 & 35.65 \\
    & GEM              & 6.50  & 1.10  & 27.30 & 81.00 & 51.60 & 44.90                & 16.90 & 32.00 & 32.66 \\
    & DFT              & 2.50  &\textbf{1.20}  & 26.60 & 78.60 & 49.60 & 41.00                & 17.80 & 30.30 & 30.95 \\
    & SED-SFT w/o mask & \textbf{13.60} & 0.60  & 31.50 & 84.00 & 54.60 & 41.00                & 21.00 & \textbf{34.50} & 35.10 \\
    & SED-SFT          & 11.20 & 0.40  & \textbf{38.40} & \textbf{85.10} & \textbf{57.70} & \textbf{53.00} &                 \textbf{22.20} &  33.70 &  \textbf{37.71} \\
  \bottomrule
\end{tabular}
}
\caption{The main results of Llama-3.2-3B-Instruction and Qwen2.5-Math-7B-Instruct across 8 math datasets. Avg@8 represents the average pass rates over eight sampling iterations for the AIME24/25 and AMC23.}
\label{tab:main_result}
\end{table*}

\section{Experiments}
\subsection{Experimental Settings}
All experiments follow a standard SFT-then-RL pipeline to evaluate whether selectively encouraging diversity during SFT improves downstream RL performance. Unless otherwise specified, we keep all training and evaluation settings identical across methods and vary only the SFT objective.

\paragraph{Backbone Models}
We conduct experiments on two instruction-tuned backbones: Qwen2.5-Math-7B-Instruct\footnote{\url{https://huggingface.co/Qwen/Qwen2.5-Math-7B-Instruct}} and Llama-3.2-3B-Instruct\footnote{\url{https://huggingface.co/meta-llama/Llama-3.2-3B-Instruct}}.

\paragraph{Datasets}
For SFT, we sample 20,000 examples from the Micomind dataset\footnote{\url{https://huggingface.co/micromind}}.
For RL, we use the Math (Level 1) training split\footnote{\url{https://huggingface.co/datasets/DigitalLearningGmbH/MATH-lighteval}}.

\paragraph{Training Details}
For SFT, we follow GEM's training setup, using a learning rate of $2\times 10^{-5}$ and DeepSpeed stage-2\footnote{\url{https://github.com/deepspeedai/DeepSpeed}}.
For RL, we apply GRPO~\cite{shao2024deepseekmath} implemented in the Verl framework\footnote{\url{https://github.com/volcengine/verl}} with batch size 256; other hyperparameters use the default GRPO configuration. We generate RL training samples by sampling each prompt 8 times with Qwen2.5-Math-7B-Instruct and filter out prompts where all samples either fail or succeed (similar to an offline DAPO procedure~\cite{yu2025dapo}), resulting in 2,069 filtered samples from an initial 5,000.

\paragraph{Compute Devices}
All SFT/RL training and evaluation are conducted on 8 NVIDIA H20 GPUs.
\subsection{Evaluation Settings}
The evaluation framework follows Qwen-2.5-Math\footnote{\url{https://github.com/QwenLM/Qwen2.5-Math}}. All of the math benchmarks are widely used in existing works, including: AIME24~\cite{aime24}, AIME25~\cite{aime25}, AMC23\footnote{\url{https://huggingface.co/datasets/math-ai/amc23}}, GSM8K~\cite{cobbe2021training}, MATH500~\cite{hendrycks2measuring}, GAOKAO-en~\cite{zhang2023evaluating}, OlympiadBench~\cite{he2024olympiadbench}, and College-MATH~\footnote{\url{https://huggingface.co/datasets/di-zhang-fdu/College_Math_Test}}, and the corresponding dataset name in the evaluation framework is: aime24, aime25, amc23, gsm8k, math, gaokao2023en, olympiadbench, and college\_math. We sample 8 times for AIME24/25 and AMC23 (the temperature is 0.7), and sample 1 time for other benchmarks (the temperature is 1.0), and then use the average pass rate as the metric.

\subsection{Baselines Selection}
We select the following representative objectives as SFT baselines: 
(1) Cross-Entropy, the most widely used training loss in SFT;
(2) GEM~\cite{li2024preserving}, which combines reverseKL and entropy loss in SFT;
(3) DFT~\cite{wu2025generalization}, which uses a confidence score to re-weight each token to debias the update.
We further include \textit{SED-SFT w/o mask} as an ablation to isolate the effect of the proposed selective masking strategy.

\subsection{Main Results}
The main results are presented in Table~\ref{tab:main_result}.
The mathematical task evaluation was conducted on two major base models, yielding the following key findings:
(1) On the base models Qwen-2.5-7B-Math and Llama-3.2-3B-Instruct, SED-SFT significantly outperformed the baseline methods. Specifically, compared to the cross-entropy baseline, SED-SFT achieved improvements of 1.20 and 2.06 points, respectively, averaged across eight widely used mathematical benchmarks.
(2) DFT and GEM, two representative SFT optimization methods, failed to maintain consistent success. DFT demonstrated a substantial advantage during the SFT phase, significantly surpassing all other methods. However, its approach greatly restricts the model's exploration space, making further improvements in the subsequent RL phase unfeasible. In contrast, GEM encourages overall diversity but ignores the token exploration space, resulting in limited adaptability to mathematical tasks.

\section{Analysis}

\subsection{Sentence-level Diversity Analysis}
To assess sentence-level diversity, we employ the Self-BLEU metric for SED-SFT and baseline methods. A lower Self-BLEU score indicates higher diversity in model-generated sentences. As shown in Table~\ref{tab:self-bleu}, both SED-SFT and GEM achieve substantially lower Self-BLEU scores compared to CE and DFT, reflecting enhanced generation diversity.

\begin{table}[ht]
\centering
\resizebox{0.5\linewidth}{!}{
\begin{tabular}{lc}
\toprule
 & Self-BLEU $\downarrow$\\
\midrule
CE & 43.12 \\
DFT & 51.26 \\
GEM & 38.53 \\
SED-SFT & 35.57 \\
\bottomrule
\end{tabular}}
\caption{The Self-BLEU of the generated results of AIME based on Llama-3.2-3B-Instruct.} 
\label{tab:self-bleu}
\end{table}

\subsection{Hyperparameter Sensitivity Study}
First, we perform parameter sensitivity examination on the ratio $r$ and the $k$ value of top-k in our method, which is shown in Table~\ref{tab:sensi_exp}. 
The results demonstrate that SED-SFT consistently outperforms cross-entropy (CE) when the ratio $r$ exceeds 0.5. Additionally, it is crucial to set the $k$ value greater than 1 to facilitate a more robust evaluation of the token exploration space.

\begin{table}[ht]
\centering
\resizebox{0.7\linewidth}{!}{
\begin{tabular}{llr}
\toprule
&Average  \\
\midrule
CE &  35.65 \\
SED-SFT($r=0.2; k=2$) & 34.18 \\
SED-SFT($r=0.5; k=2$) & 35.73 \\
SED-SFT($r=0.7; k=2$) & 37.71 \\
SED-SFT($r=0.8; k=2$) & 35.68 \\
SED-SFT($r=0.7; k=1$) & 34.68 \\
SED-SFT($r=0.7; k=3$) & 36.89 \\
\bottomrule
\end{tabular}}
\caption{Sensitivity test on the hyperparameters.} 
\label{tab:sensi_exp}
\end{table}

\section{Related Work}
Prior work has explored improving exploration efficiency mainly at the RL stage~\cite{wang2025beyond, cui2025entropy, zhu2025proximal}. These approaches typically rely on entropy-based control during policy updates, and are therefore not directly applicable to the SFT stage, where mode collapse is induced by token-level cross-entropy fitting.
To contextualize SED-SFT, we categorize related SFT optimization methods into three streams:
\paragraph{RL Policy Integration} DFT~\cite{wu2025generalization} incorporates RL policy-update ideas into SFT by preventing overly large gradients during model updates to preserve the model's exploration capacity. ASFT~\cite{zhu2025anchored} further improves the performance based on DFT by solving the bias shifting problem. However, ASFT has a much higher computational cost for inducing a reference model. While this line adjusts the update direction, it does not fundamentally increase diversity and can reduce exploration space, thereby limiting subsequent RL improvement.

\paragraph{Diversity Modeling}
GEM~\cite{li2024preserving} explicitly integrates diversity into the objective function to directly encourage varied generation. However, this approach overlooks the distinct nature of different tokens, which can potentially lead to degradation in areas where accuracy should be prioritized. Consequently, they lack robustness in tasks with high accuracy requirements ({\em e.g.}, mathematics), leading to decreased model performance.

\paragraph{Selective Gradient Updating}
Selective updating strategies, such as CTF~\cite{ruan2025enhancing} require pre-determining the important update locations, which typically demands favorable model characteristics, high-cost algorithms, or prior knowledge. This results in extremely high inconsistency and poor generalizability for token selection algorithms in general models~\cite{ming2025one,kim2025sft}.

\section{Conclusion}
We propose SED-SFT, a method that utilizes entropy information to enhance supervised fine-tuning by incorporating a simple entropy regularization function and a masking strategy quantifying the token exploration space using cumulative top-k token probability. Across experiments on Llama-3.2-3B-Instruct and Qwen2.5-Math-7B-Instruct and 8 mathematics benchmarks in the SFT-then-RL paradigm, the experimental results demonstrate that SED-SFT consistently enhances the model's generation diversity and consistently improves performance after RL. Meanwhile, according to the analysis model generation result, SED-SFT increases the sentence-level diversity significantly.




\bibliography{custom}

\appendix


\section{Prediction Probability Heatmap}\label{sec:prob_heatmap}
\begin{figure*}[ht]
    \centering
    \includegraphics[width=\linewidth]{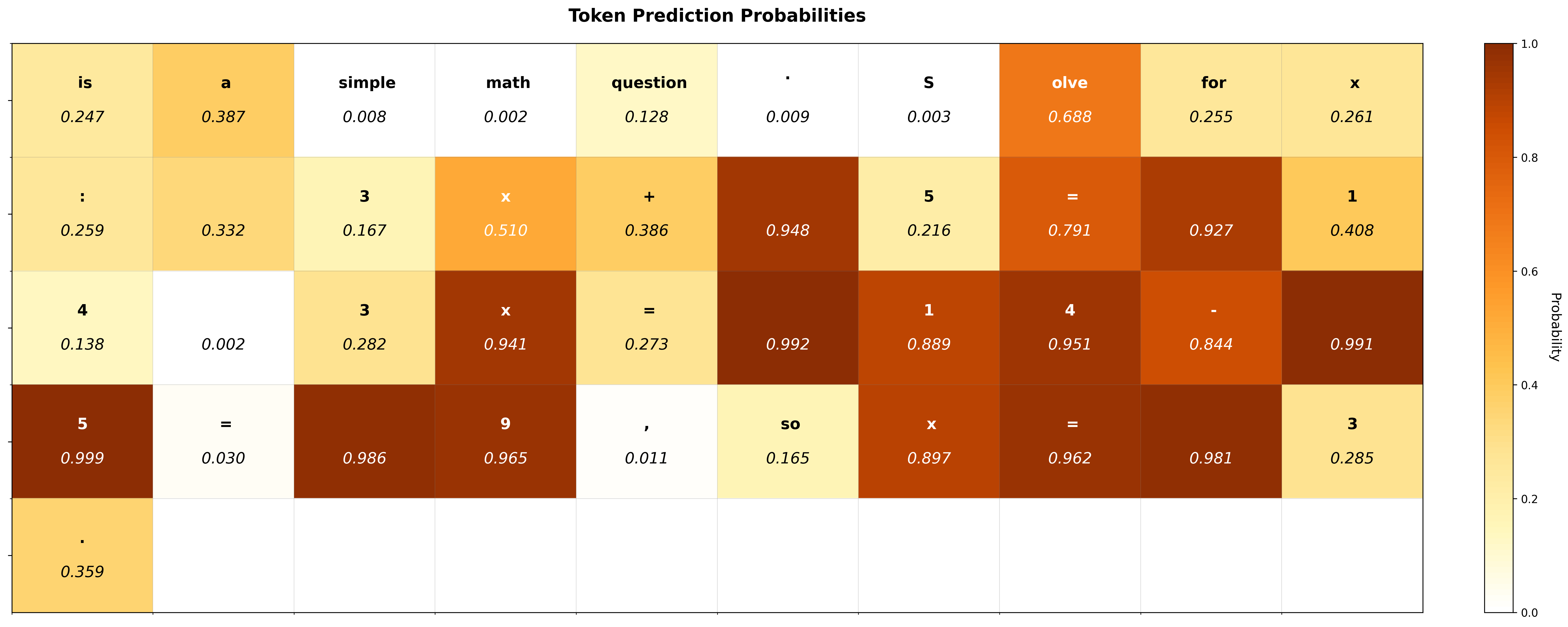}
    \caption{The heatmap for the probability of the labels on each position. The case is a simple math problem and evaluated on Qwen-0.5B}
    \label{fig:heatmap_example}
\end{figure*}

 As shown in Figure~\ref{fig:heatmap_example}, we visualize the probability distribution of labels at each position using heatmaps to demonstrate that tokens with exceptionally high probabilities typically correspond to those with limited exploration space.

\section{Heatmap on the statistic of the cumulative probability of top-k}\label{sec:divergence_heatmap}
\begin{figure*}
    \centering
    \includegraphics[width=\linewidth]{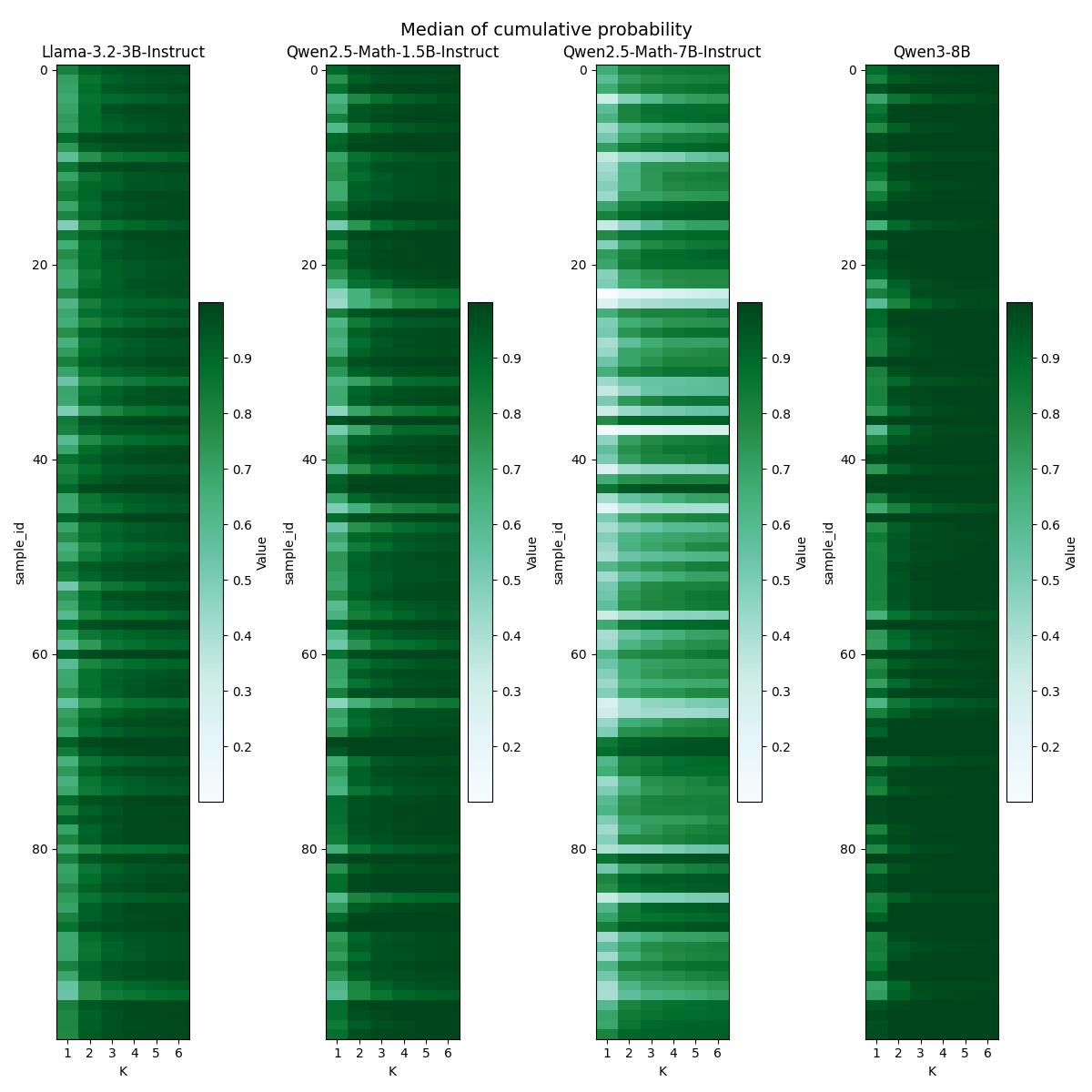} 
    \caption{The visualization of the cumulative probability with different k values in Top-k.}
    \label{fig:cumsum_divergence}
\end{figure*}

We sample 100 math task examples from the OpenR1\_Math~\footnote{\url{https://huggingface.co/datasets/open-r1/OpenR1-Math-220k}} dataset and obtain the cumulative probability for different k values across four representative backbone models: Llama-3.2-3B-Instruct, Qwen2.5-Math-1.5B-Instruct, Qwen2.5-Math-7B-Instruct, and Qwen3-8B. As shown in Figure~\ref{fig:cumsum_divergence}.  As illustrated in Figure\ref{fig:cumsum_divergence}, the results reveal significant divergence in cumulative probability distributions across different models and samples.
\label{sec:appendix}

\end{document}